\documentclass[pdflatex,sn-mathphys-num]{sn-jnl}

\usepackage[T1]{fontenc}
\usepackage[utf8]{inputenc} 
\usepackage{kpfonts}
\AtBeginDocument{\selectfont}

\usepackage{graphicx}
\usepackage{multirow}
\usepackage{amsmath,amssymb,amsfonts}
\usepackage{amsthm}
\usepackage{mathrsfs}
\usepackage[title]{appendix}
\usepackage{xcolor}
\usepackage{textcomp}
\usepackage{manyfoot}
\usepackage{booktabs}
\usepackage{algorithm}
\usepackage{algorithmicx}
\usepackage{algpseudocode}
\usepackage{listings}
\usepackage{subcaption}
\usepackage{bm}

\theoremstyle{thmstyleone}

\theoremstyle{thmstyletwo}

\theoremstyle{thmstylethree}

\begin{document}

\title[Universality in Collective Intelligence on the Rubik's Cube]{Universality in Collective Intelligence \\ on the Rubik's Cube}

\author*[1]{\fnm{David} \sur{Krakauer}}\email{dk@santafe.edu}
\author[1,2]{\fnm{G\"ulce} \sur{Karde\c{s}}}\email{gulcekardes@gmail.com}
\author[2,3]{\fnm{Joshua A.} \sur{Grochow}}\email{jgrochow@colorado.edu}

\affil*[1]{\orgname{Santa Fe Institute}, \orgaddress{\street{1399 Hyde Park Road}, \city{Santa Fe}, \postcode{87501}, \state{NM}, \country{USA}}}

\affil[2]{\orgdiv{Department of Computer Science}, \orgname{University of Colorado Boulder}, \orgaddress{\street{1111 Engineering Dr., ECOT 717, 430 UCB}, \city{Boulder}, \postcode{80309}, \state{CO}, \country{USA}}}

\affil[3]{\orgdiv{Department of Mathematics}, \orgname{University of Colorado Boulder}, \orgaddress{\street{Campus Box 395}, \city{Boulder}, \postcode{80309}, \state{CO}, \country{USA}}}

\abstract{
Progress in understanding expert performance is limited by the scarcity of quantitative data on long-term knowledge acquisition and deployment. Here we use the Rubik’s Cube as a cognitive “model system” existing at the intersection of puzzle solving, skill learning, expert knowledge, cultural transmission, and group theory. By studying competitive cube communities, we find evidence for universality in the collective learning of the Rubik’s Cube in both sighted and blindfolded conditions: expert performance follows exponential progress curves whose parameters reflect the delayed acquisition of algorithms that shorten solution paths. Blindfold solves form a distinct problem class from sighted solves and are constrained not only by expert knowledge but also by the skill improvements required to overcome short-term memory bottlenecks, a constraint shared with blindfold chess. Cognitive artifacts such as the Rubik’s Cube help solvers navigate an otherwise enormous mathematical state space. In doing so, they sustain collective intelligence by integrating communal knowledge stores with individual expertise and skill, illustrating how expertise can, in practice, continue to deepen over the course of a single lifetime.
}


\maketitle
\section{Introduction}\label{sec1}

The Rubik's Cube, invented by Erno Rubik in 1974, is a popular puzzle and the physical embodiment of a mathematical structure, the Cayley graph of a permutation group generated by face turns \cite{cooperman1990applications}. Solving the puzzle involves a sequence of moves forming a path in its state graph from a scrambled state to the solved state, where each edge is a legal move. In the solved state, each of the six faces of the $n \times n \times n$ cube is monochromatic. The longest path through this graph from a maximally jumbled start to the solution (i.e., the diameter of the graph) is given by its ``God's Number'' \cite{rokicki2014thirty} (${G}_n$), which for the 3-cube is $20$: every position is solvable in at most 20 face turns (counting both quarter-turns and half-turns as single moves), and some positions require exactly 20. 

The mapping of a popular puzzle directly onto a formal mathematical structure offers numerous advantages, whereby formal concepts like entropy, difficulty, search, and optimality, as well as cognitive concepts such as skill and expertise, can be formally connected through a cognitive artifact \cite{rothstein2019shape,danesi2020anthropology}. We describe the simplification (demonstrable reduction in dimension) of combinatorial search problems by physical artifacts as the \emph{Principle of Materiality}, and show how the application of memorized algorithms to the Rubik's Cube reveals fundamental characteristics of collectively intelligent systems \cite{KardesKrakauerGrochow2025PCoCA}. The Rubik's Cube has provided the basis for an ongoing series of tournaments in which highly skilled competitors seek to minimize both the time and the number of steps required to solve sighted and blindfolded $n$-cubes, where $n$ denotes the cube's linear dimension \cite{scheffler2017cracking}. In this paper, we analyze record-breaking solutions for all sighted cubes from $n=[3,7]$ spanning up to 19 years of competitive play and blindfolded competition with $n_b=[3,5]$ spanning 17 years. These are recorded by the World Cube Association---a volunteer-organized association that has overseen thousands of competitions and millions of recorded solves under refereed conditions (\url{https://www.worldcubeassociation.org}). 

\subsection{Algorithms} \label{sec3}

Efficient human solutions of the Rubik’s Cube rely on memorized “algorithms”, pattern–action rules that map a perceived global or local cube configuration to a specific sequence of face turns \cite{harris2008speedsolving,scheffler2017cracking}. The shared repertoire of these algorithms in the speedcubing community has grown steadily over the past several decades. From the 2000s onward, CFOP (the Fridrich method) became the dominant method in speed-solving competitions. In its fully algorithmic form, CFOP requires on the order of a hundred memorized algorithms for the first two layers and the last layer, although many solvers rely on reduced two-look variants with only a few dozen core cases. Additional methods subsequently gained popularity, including Roux, Petrus, and ZZ \cite{duberg2015comparison}, which are well known within the speedcubing community. Together with specialized last-layer sets such as ZBLL, which alone comprises hundreds of distinct cases, these methods have expanded the shared algorithmic repertoire to several hundred commonly used patterns. Online resources maintained by the community now catalogue thousands of distinct move sequences used for speed-solving the cube \cite{rubik2024rubik}. Many of these are committed to memory and practiced to a high level of skill. 

The cube is also solved in dedicated blindfolded competitions. After an initial inspection phase, during which solvers commit the scrambled cube to memory using memory-palace techniques \cite{yates2013art,carruthers1992book}, they solve it while blindfolded. During memorization, each target sticker (edge or corner) is assigned a letter; letter pairs are then turned into images or words and placed along a route in a mental “palace” to encode the solution sequence. To accommodate this heavy mnemonic load, blindfolded solvers typically rely on a smaller, more standardized repertoire of algorithms.

\begin{figure}[t]
  \centering
  \includegraphics[width=\columnwidth]{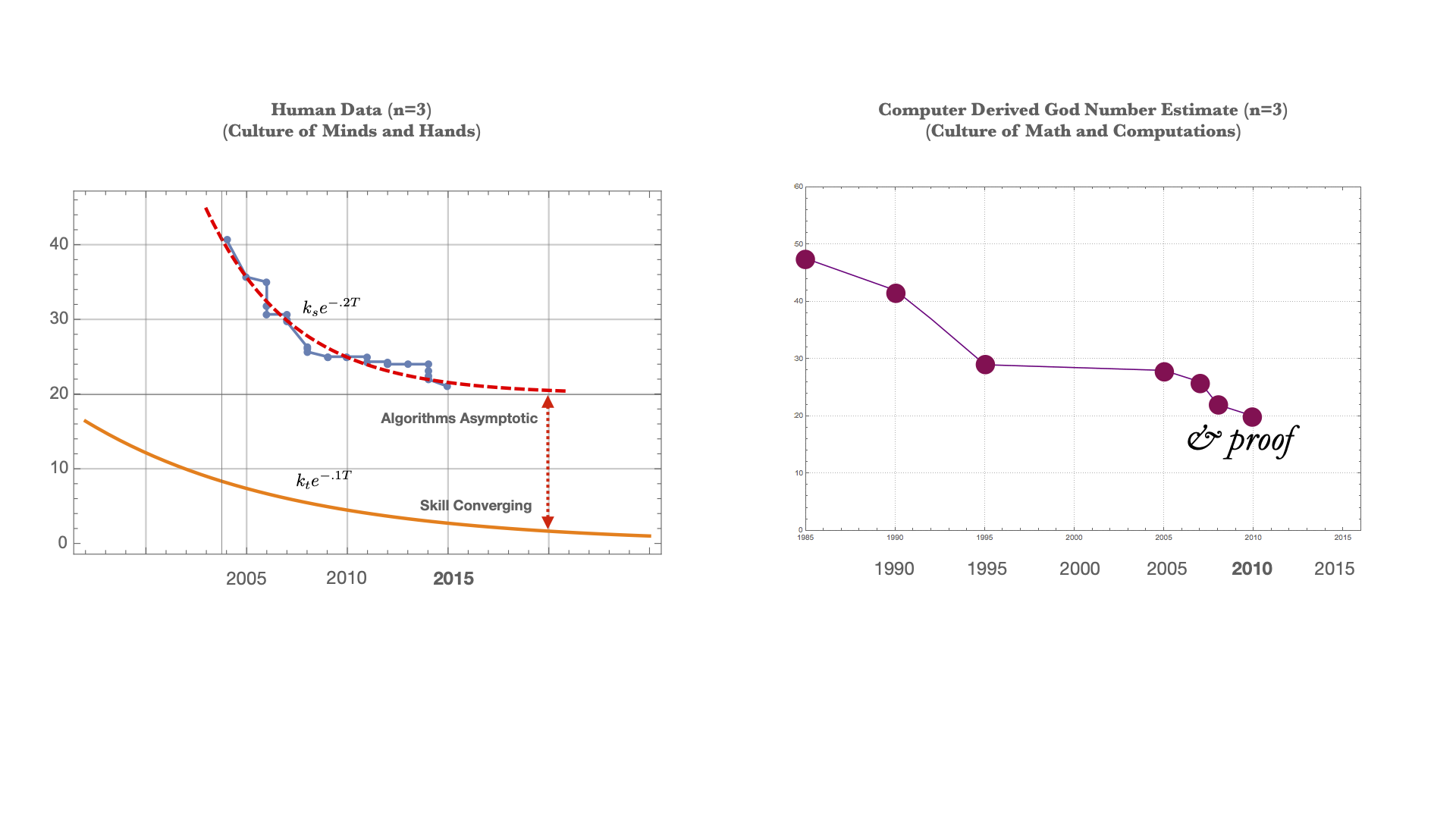}%
  \caption{%
   (A) Progress curves for the 3-cube showing the average number of twists required to solve the cube in a minimal number of face turns in competition (blue points and lines, with exponential fit superimposed in red) and the associated average solution times (orange). The rate of improvement in face turns is twice that of solution time, illustrating the kinematic or motor constraints associated with executing optimal algorithmic procedures. (B) Progress curves in estimating the God's Number for the 3-cube showing the average number of twists required to solve the cube optimally as estimated by numerical computation. In 2005, competitors and computers were in rough agreement about the fewest-move-count (FMC) solution. By 2010, large-scale computer calculations had established the God's Number as 20 moves, a bound that top competitors first matched in official fewest-moves events around 2015.
  }
  \label{fig:f2ab}
\end{figure}

\subsection{Formalizing the State Space} \label{sec2}

The Rubik's Cube has on the order of $4.3 \times 10^{19}$ reachable states, which makes its state graph astronomically large. In practical terms, there is no way to explore all configurations by brute force. The cube’s state space is the Cayley graph of its group $G$ under a chosen move metric with a generating set. Because this graph is vertex-transitive, local neighborhoods look statistically similar across configurations; \emph{absent global structure or learned algorithms}, a solver that sees only immediate neighbors has negligible distance signal from the solution. Hence local moves are nearly symmetric with respect to distance, and naive locality does not guarantee monotonic progress towards the solution. By 2010, computer simulations had determined the God's Number to be 20, which was matched by competitors around 5 years later in 2015. 

We work in the Cayley graph of the cube group $G$ generated by a fixed move set $M$; 
throughout we take $M$ to be the half–turn metric (HTM). 
For a cube state $g \in G$ let $\ell_M(g)$ denote its distance from the solved state, 
i.e., the length of a shortest word in the generators $M$ that takes the solved cube to $g$.
For each radius $r \ge 0$ we define the ball
$
  B_M(r) \;:=\; \{\, g \in G : \ell_M(g) \le r \,\},
$
and write
$
  \gamma_M(r) \;:=\; |B_M(r)|
$
for the number of states in this ball.  
Our (accumulated, or ball) entropy at radius $r$ is then
 $H_M(r) \;:=\; \log_2 \gamma_M(r).$
We also consider the shell at radius $r$,
  $\Sigma_M(r) \;:=\; \{\, g \in G : \ell_M(g) = r \,\},$
with shell size $
  S_M(r) \;:=\; |\Sigma_M(r)|,
$
noting that $S_M(0) = 1$ for the unique solved state. For $r \ge 1$ we define the (average) branching factor
$b_M(r) := S_M(r)/S_M(r-1)$.

\begin{figure}[t]
  \centering
  \includegraphics[width=0.95\columnwidth]{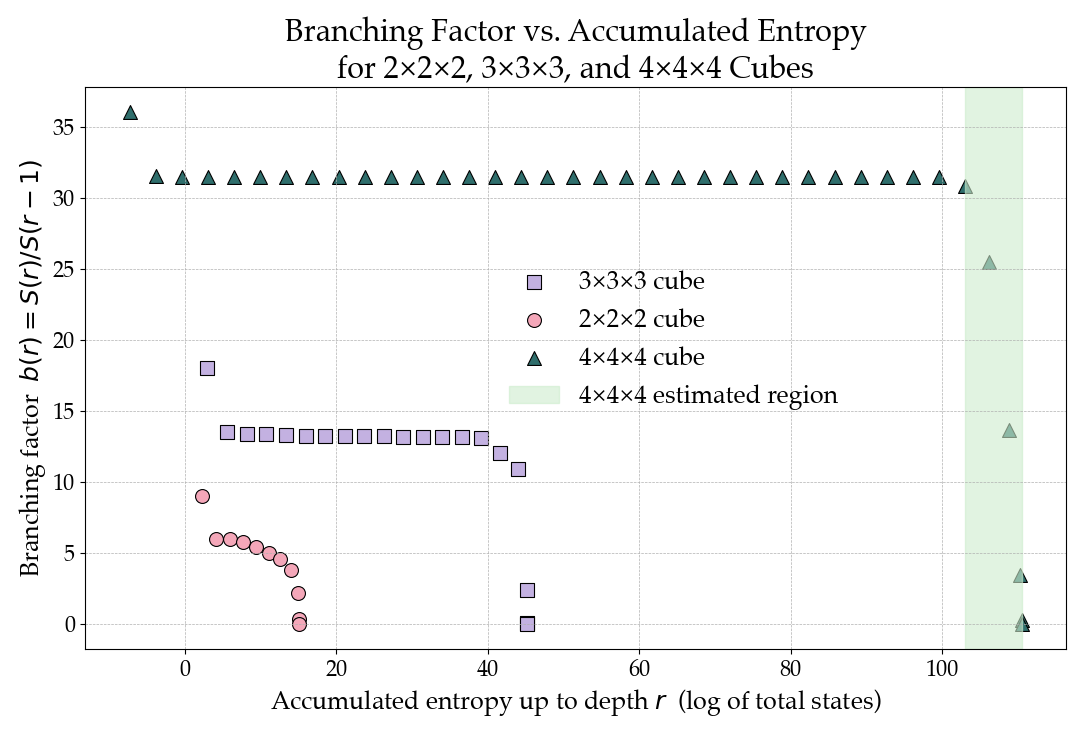}%
  \caption{%
   Increasing the linear size of the cube raises both the plateau branching factor (effective decision-tree branching) and the slope of the entropy-radius relation, leading to much faster growth of the reachable state space and hence more rapid “shuffling”.
  }
  \label{fig:branching}
\end{figure}

Figure~\ref{fig:branching} shows the average branching factor and the corresponding entropy at each depth for three cube variants: the Pocket Cube $(2\times2\times2)$, the standard Rubik’s Cube $(3\times3\times3)$, and the $4\times4\times4$ Cube \footnote{Estimates; see the code implementing these computations 
\href{https://github.com/gulcekardes/Rubik-Cube-Artifact}{Rubik's Artifact}.}. Larger cubes expand faster and for longer: their branching factors are higher and stay high deeper into the state graph, so the number of configurations within radius $r$ and therefore the entropy grows exponentially more quickly. They are increasingly more difficult with exponentially larger state spaces and local expansion.

On a state graph where $b(r)$ plateaus at a value $>1$, the ball size satisfies $\gamma_{\mathrm{M}}(r)\asymp c^{\,r}$ for some constant $c>1$, so $H_{\mathrm{M}}(r)=\log_2\gamma_{\mathrm{M}}(r)$ is asymptotically linear in $r$. Consequently, any search that expands nodes without strong global heuristics must examine on the order of $\gamma_{\mathrm{M}}(r)$ states to depth $r$ (exponential in $r$). Likewise, a myopic policy with forward-move probability $p_f$ behaves like a biased 1-D walk with drift $\mu=2p_f-1$, giving expected solve length $\approx r/\mu$ from distance $r$. Effective algorithms therefore either (i) reduce the branching, (ii) reduce the effective depth, or (iii) introduce shortcut macros that increase the forward drift $\mu$ \cite{KardesKrakauerGrochow2025PCoCA}. In all cases, performance gains correspond to lowering the entropy slope or increasing the drift, which is a linkage that we exploit when inferring the universal learning curves from fits of progress curves.

\subsection{Universality} \label{sec4}

In Figure~\ref{fig:f2ab}A we plot annually averaged record-breaking solution times (progress curves) and solution path lengths for the 3-cube. Solution steps are plotted in total half turns of the cube starting from the shuffled initial condition and culminating in the solved configuration. The corresponding solution time is plotted in seconds. 

Path length is discrete and shows an episodic pattern of reduction with an upper bound determined by the worst-case performance when following the optimal solution policy. This is equal to the diameter of the Cayley graph and is described as the God's Number, ${G}_3=20$. Solution time is continuous and shows signs of reaching an asymptotic performance between 3–4 s. 

The best-fitting functions for both the path-length and solution-time data are exponential functions: $k_p e^{\frac{1}{5}T}$ and $k_t e^{\frac{1}{10}T}$, where $k_p$ is the initial path length and $k_t$ the solution time at the start of recorded competitions. The exponential rate of improvement for path length is twice that of the solution time.

In Figure~\ref{fig:f2ab}B we chart the literature’s computational estimates of the 3-cube God's Number across the corresponding years \cite{rokicki2014thirty}. Up until 2005, the computational estimate and human performance were approximately equal. By 2010, the God's Number had been settled through computation and algebraic arguments to be 20; a bound that began to be reached in official fewest-moves events around 2015. This is a nice illustration of parallel progress between cube theorists and cube practitioners. 

\begin{figure}[t]
  \centering
  \includegraphics[width=\columnwidth]{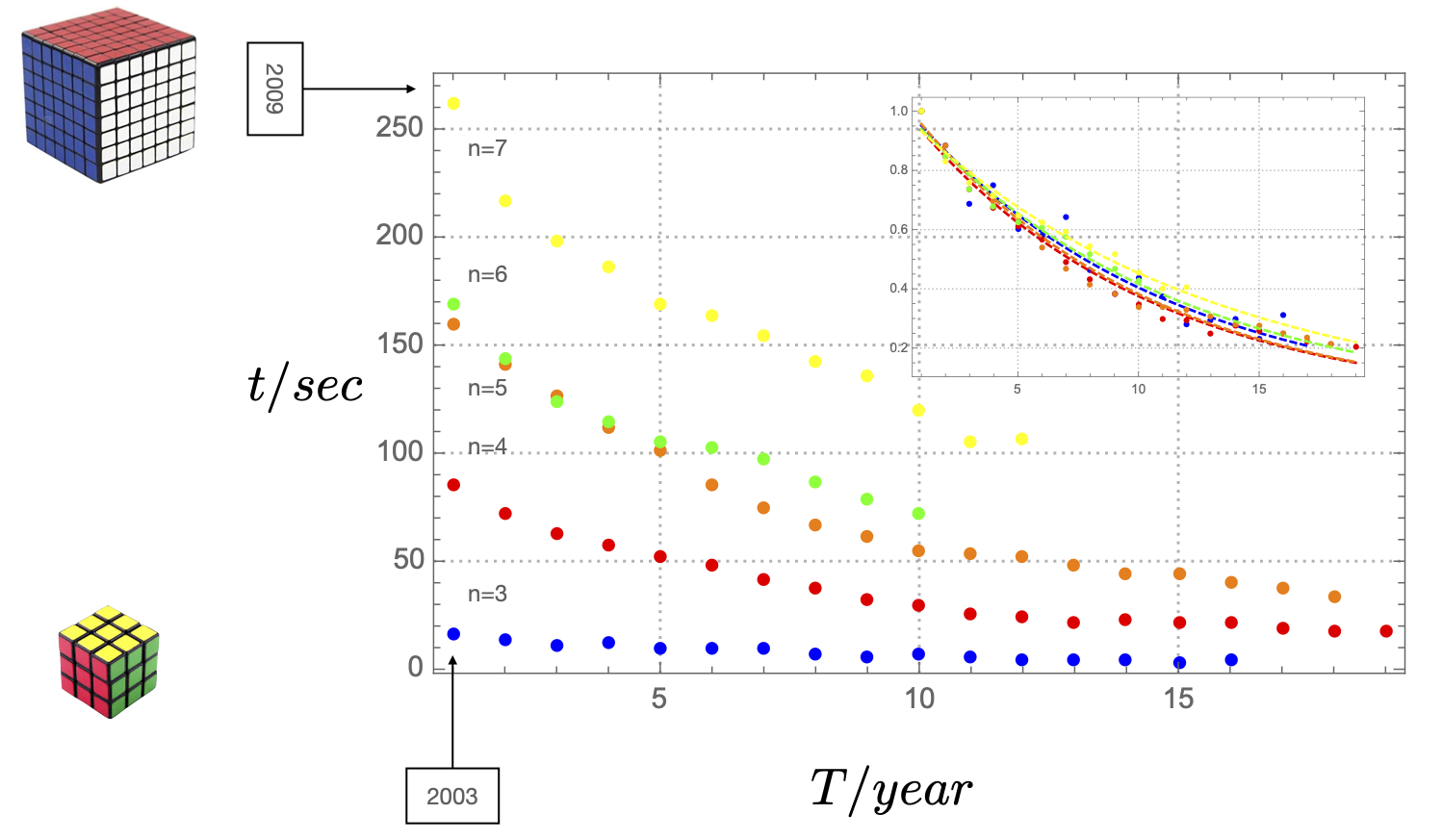}%
  \caption{%
   Progress data derived from an archive of record-breaking competitions spanning 20 years. The start of competition for each cube is indicated as year~1. Year~1 for the 3-cube was 2003 and year~1 for the 7-cube 2009. The inset shows the collapse of all five progress curves onto a single universal curve by normalizing each curve by its maximum. All curves are revealed to be the same curve differing only in their asymptotic value (translational offset in time) determined by the size of the cube. Thus there is no evidence that competitors are more skilled on smaller cubes. 
  }
  \label{fig:f3}
\end{figure}

In Figure~\ref{fig:f3}, the annually averaged record-breaking solution times are plotted in seconds for five different cubes spanning $n=3$ to $n=7$. These are described as progress curves. The chronological start date for each cube is different; hence time is shown counting from its first year in recorded competition. Indicated in parentheses are the total number of record breakers for each cube's complete progress curve. For all cubes one observes an identical exponential improvement in solution time. The exponential is by far the best-fitting two-parameter family of functions: the Bayesian information criterion (BIC) differences satisfy
$
\mathrm{BIC}_{\text{linear}} - \mathrm{BIC}_{\exp} > 30
\,\text{and}\,
\mathrm{BIC}_{\text{power}} - \mathrm{BIC}_{\exp} > 10.
$. The inset figure shows the progress curves for all $n$-cubes when rescaled by the reciprocal of their maximum (first competitive solution time) $k_{t0}$, illustrating collapse onto a single universal scaling curve. Regardless of $n$, the number of different competitors contributing to any annual collective solution, or the chronological year, we find that solution times halve on average every ten years. The progress-curve half-life of all $n$-cubes appears to be a stylized cognitive fact for this puzzle. 

\subsection{Mechanisms} \label{sec5}

\begin{figure}[t]
  \centering
  \includegraphics[width=\columnwidth]{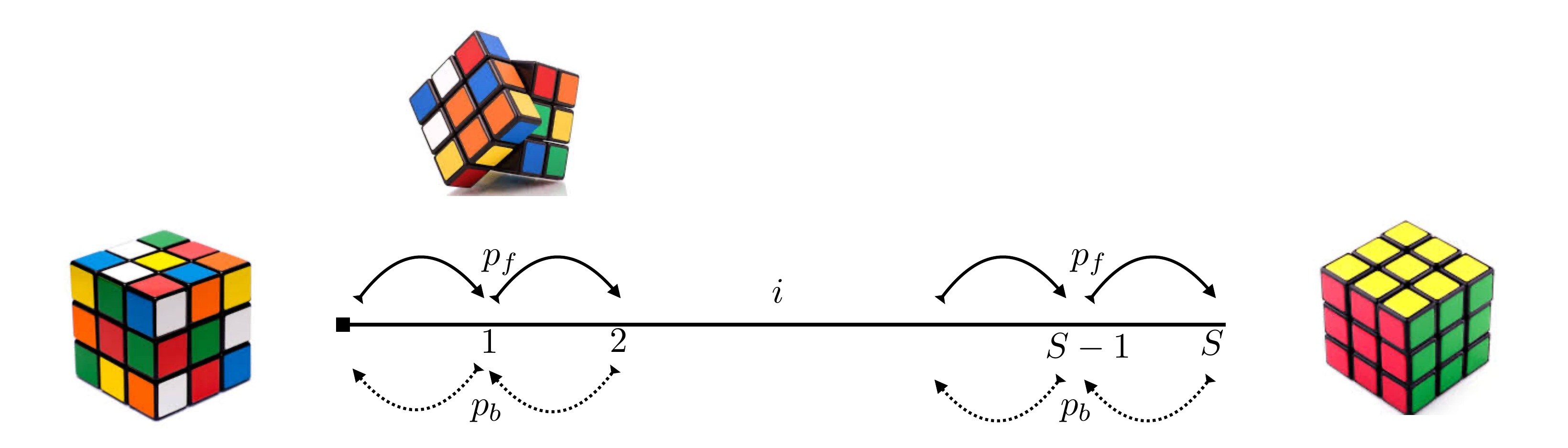}%
  \caption{%
   The solution of a cube can be approximated by a discrete random walk through a graph of (approximately) constant branching factor (the same number of moves available at every step), where the accuracy of a solve is captured by a forward probability, or drift, $p_f >0.5$. When $p_f = 1$ the cube will be solved in at most the God's Number of steps. With less than optimal use of algorithms, $p_f<1$, the cube is solved in more steps. The very high-dimensional nature of the Cayley graph is summarized in the value of $p_f$, which is an expertise metric for solving the cube. 
  }
  \label{fig:f4}
\end{figure}

We analyze the universal exponential progress by treating solving as a first-passage problem on a near-regular (with respect to average branching factor) state graph and by modeling the community's algorithm discovery and adoption as a collective-intelligence process (Figure~\ref{fig:f4}). 

Any turn of a single face of the cube corresponds to a generator of the Cayley graph. From a state at distance $r>0$, one move (the inverse of the last) typically backtracks to $r-1$, while most admissible moves lead to $r+1$ (up to short relations). Thus an uninformed random move has an outward bias and, on average, increases distance—except near the diameter band where saturation effects appear. “Optimal” moves increase the probability of stepping inward. Suboptimal turns make the trajectory more diffusive with frequent excursions away from the solution. The expected first-passage time for a competitive cuber on the 3-cube is proportional to $G_3/(p_f-p_b)$, where $(p_f+p_b) = 1$. A perfect solve ($p_f=1$) is ballistic and takes at most $G_3=20$ moves (HTM). The parameter $p_f$ captures expertise; for competitive solvers $p_f>\frac{1}{2}$.

The value of $p_f$ in any given year, which influences forward drift through the cube, is derived from the average expertise of the pooled best competitors, and encodes their current knowledge and skill at solving the cube. In Figure~\ref{fig:f2ab} we were able to break down human performance into the number of turns required to solve the cube in a year (a measure of expertise that captures knowledge of algorithms) and the solution time, which includes the skill with which the cube is solved. We would like to be able to distinguish elements of expertise from competitor skill based entirely on the competitive progress curves plotted in Figure~\ref{fig:f3}. In order to approximate this separation we relate the progress curves to ``latent learning curves''—curves that we infer from knowledge of the Rubik's cube state space using the recorded progress curves. 

Learning across a variety of different domains can be captured with a sigmoidal function. A sigmoidal curve results from a variety of learning processes including Bayesian updating, reinforcement learning, and operant conditioning, and can be viewed as an idealized form of many learning curves \cite{murre2014s}. We operationalize $p_f$ as a learning curve with shape parameters $r$ and $\tau$, where $r$ is the learning rate and $\tau$ the inflection point. Hence,
\begin{equation}
    p_f(T) = \frac{1}{2} + \frac{1}{2(1+e^{r(\tau-T)})}.
\end{equation}
When substituted into a first-passage-time approximation, this yields a progress curve of the form
\begin{equation}
 P_c(T) = G_n\bigl(1+e^{r(\tau-T)}\bigr).   
\end{equation}

The learning rate $r$ is a proxy for the skill of a cuber and is a measure, for a given level of knowledge, of how well a competitor executes a memorized set of algorithms on a given $n$-cube. The inflection parameter $\tau$ is a proxy for expertise, since it establishes, for a given level of skill, how quickly the Cayley graph is traversed with a given number of short-cut algorithms. One can also think of $\tau$ as capturing the time required to accumulate the requisite number of short cuts to efficiently solve the cube. For the small range $3 \le n \le 7$ we consider here, the available bounds
indicate that $G_n$ remains $\Theta(1)$ in absolute scale (on the order of a few
tens of moves), so we normalize $G_n$ to $1$ in our fits. Note that in this approximation the God's Number does not influence $r$ or $\tau$ but only the intercept on the $y$-axis. By fixing the God's Number we are forcing the combinatorial aspects of increasingly large cubes outside of the diameter of the Cayley graph into the inflection point $\tau$.  

\begin{figure}[t]
  \centering
  \includegraphics[width=0.75\columnwidth]{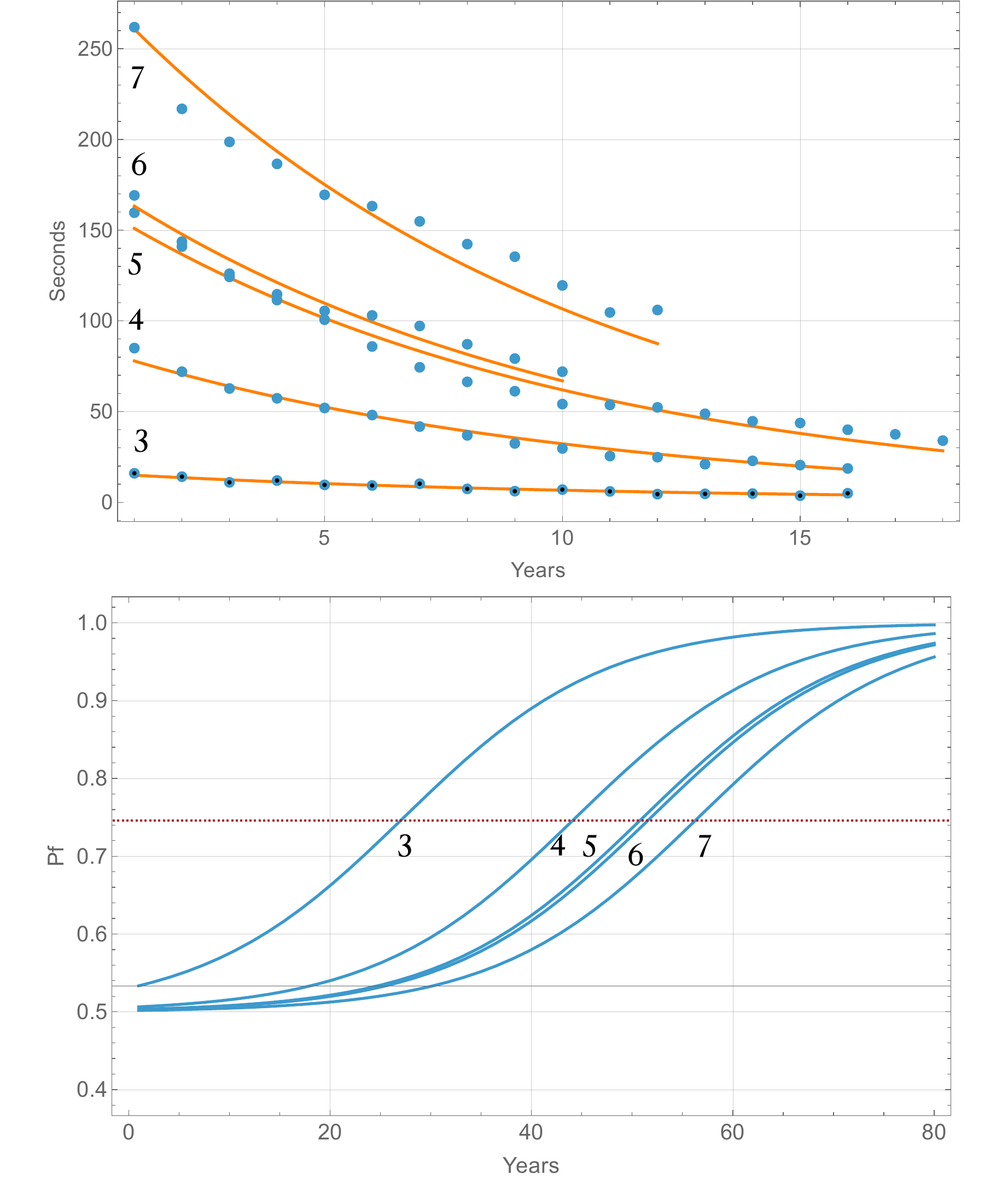}%
  \caption{%
  Collective progress curves and learning curves for Rubik's Cube competitions spanning 20 years. (A) Competitive data and optimal fits to the five $n$-cubes using Eq.~2. (B) The solution to Eq.~1 based on fits to Eq.~2. The learning rates across all cubes are identical; they differ only in their initial $p_f$ values and in their inflection points. Larger cubes show significant delays in the acquisition of expertise required to rapidly solve the cube. 
  }
  \label{fig:f5}
\end{figure}

In Figure~\ref{fig:f5}A we plot progress curves for the five cubes based on Eq.~2 and in Figure~\ref{fig:f5}B their associated learning curves based on Eq.~1. As per the universality of exponentials in Figure~\ref{fig:f3}, the learning rate $r$ is constant across all cubes. The cubes only differ in the value of $\tau$, which increases with the value of $n$ for each larger $n$-cube. The larger the cube, the longer it takes at a given fixed level of skill to significantly increase $p_f$. There is no evident pattern of geometric scaling of $\tau$ as a function of $n$ in the recorded competitions. There is a clustering of the 4–7-cubes above the 3-cube. None of the cubes have yet reached their projected maximum level of expertise and only the 3-cube is reaching its inflection point. 

\begin{figure}[t]
  \centering
  \includegraphics[width=0.85\columnwidth]{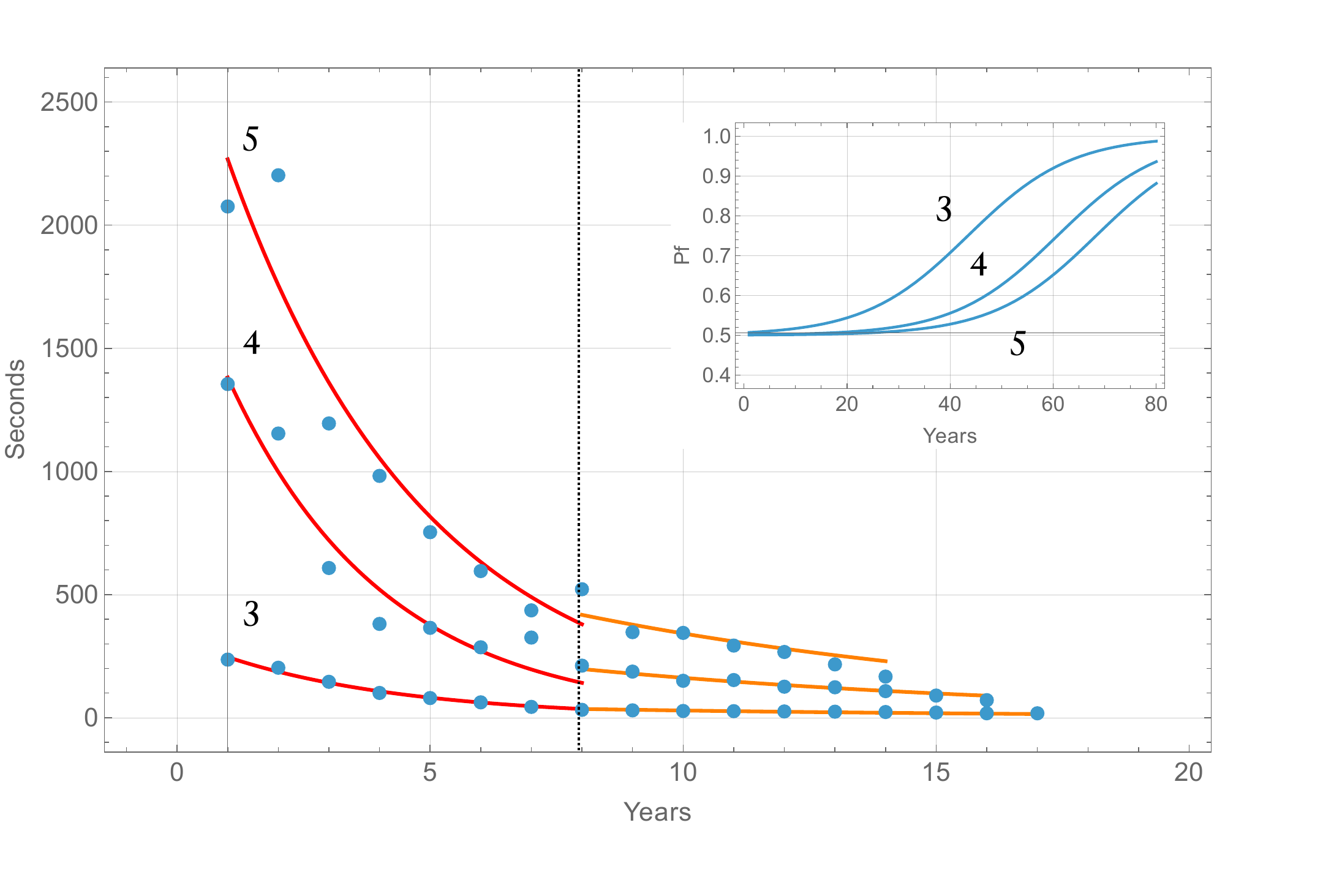}%
  \caption{%
    Collective progress curves and learning curves (inset, blue) for blindfolded Rubik's Cube competitions spanning 20 years. As in Figure~\ref{fig:f5}, the competitive data are plotted together with optimal fits to the three $n_b$-cubes using Eq.~2, and Eq.~1 is then used to derive the learning curves. In the first 8 years of competition the learning rates across all $n_b$-cubes are identical and they differ only in their initial $p_f$ values (red lines). After 8 years the $n_b$-cube curves collapse onto their corresponding sighted $n$-cube curves (orange lines). Solving the cubes blindfolded shows longer delays than the sighted cubes, reflecting an initial phase of learning memory-palace techniques and a second phase corresponding to sighted solves. 
  }
  \label{fig:f6}
\end{figure}

In Figure~\ref{fig:f6} we plot three cubes $n_b=3$–5 used in blindfold competition. For the first 8 years of competition, all of the cubes are described by a single exponential curve, with learning rate parameter $r=0.3$, which is greater than all the sighted curves. After 8 years of competition, the three blindfold cubes collapse onto the universal curve of the sighted cubes with learning rate $r=0.1$. The 8-year segmentation is determined by change-point estimation—the time at which the time series is bisected by a significant change in its variance \cite{sun2024snseg}. Blindfold inflection points are delayed beyond those of sighted cubes, with the 3b-cube taking as long as the 4-cube to reach its inflection point. The 4–5b-cubes take longer to reach inflection than all sighted cubes. Thus the blindfolded cube shows a marked biphasic progress-curve profile that is independent of the difficulty of solving any given cube and exhibits two exponential universality classes. We might say that the blindfolded solve is more skillful (faster rate of learning) in its early execution but less expert (fewer algorithms acquired) over the complete course of competition. 

\begin{figure}[h!]
  \centering
  \includegraphics[width=0.85\columnwidth]{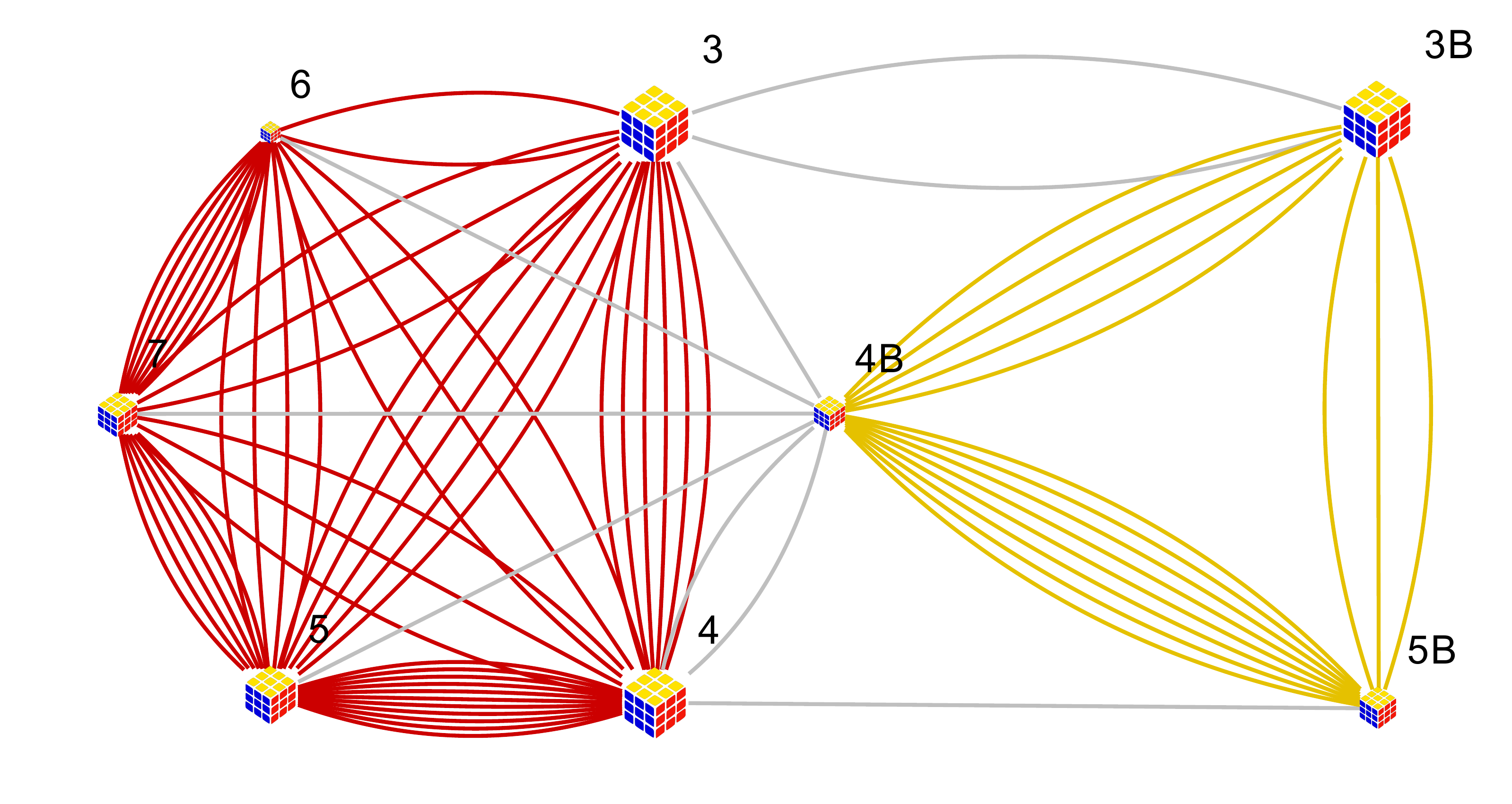}%
  \caption{%
   The social network of competitive, record-holding Rubik's Cube solvers. Each node is a set of record-breaking competitors spanning up to 20 years of competition. The size of the node is proportional to the logarithm of the number of competitors on that size cube. Each edge describes a single competitor setting a record on two connected cubes. The graph has been color coded by a community-structure detection algorithm. The community of sighted solvers is connected by red edges. The community of blindfolded solvers is connected by yellow edges. The two communities are connected by gray edges. 
  }
  \label{fig:f7}
\end{figure}

In Figure~\ref{fig:f7} we show the “social network” of competitive cubers contributing to the data. Each node is an $n$-cube (red subgraph) or $n_b$-cube (yellow subgraph). Two nodes are connected when a single competitor participates on both cubes. Hence the sighted 3-cube and sighted 4-cube will be connected by an edge when one or more competitor competes on both Rubik's Cubes. The social networks within sighted and blindfolded communities are fully connected (at least one competitor for each cube competes on all other cubes), whereas the two communities are not fully connected. Interestingly, members of the set of competitors on the 4b-cube compete on all sighted cubes, whereas competitors on the 3b-cube only compete on the 3-cube and competitors on the 5b-cube compete only on the 4-cube. The modularity of the graph is derived from community-structure detection based solely on connectivity \cite{newman2006modularity} and cleanly separates the sighted and blindfolded communities, color-coded into red and yellow edges.

\section{Discussion}\label{sec6}

For the generalized $n \times n \times n$ cube, the decision problem “is there a solution in $\leq k$ slice turns?” is NP-complete. Separately, the worst-case optimal move count (God's Number) scales as $\Theta\!\left(n^2 / \log n\right)$ under standard slice-turn metrics \cite{demaine2017solving}. Thus the number of moves required in the worst case increases superlinearly with $n$ (note that this is a property of the state graph, not a consequence of NP-completeness). Larger cubes have exponentially larger state spaces and a superlinearly growing worst-case solve length; hence they are theoretically harder (from a worst-case/optimality standpoint). We show that this is not the case for human solvers when deploying suitable heuristics. In fact, not only does the cube not get harder as $n$ increases, it remains the same puzzle instrumentally, with the same learning rate across all cubes, as shown by the collapse of progress curves onto one universal function. This universality reflects the power of approximate, algorithmic strategies grounded in the solver’s perceptual and motor interaction with the physical cube.

Previously, we found that in situations that promote collective intelligence, competition leads to significant increases in problem-solving accuracy \cite{brush2018conflicts}. This effect is facilitated by the global information provided by a physical artifact, which complements human cognition by exploiting the modalities of touch and vision \cite{Hutchins2014-qa,Clark1998-vi,krakauer2016will}. 

These findings make ideas from computational complexity theory relevant to cognition by shifting the focus from worst-case optimality to a distributional, advice-aided regime in which practical hardness is governed by two graph parameters: depth scale and advice-induced forward drift. In our setting, human algorithm learning is better modeled as augmenting the state graph with learned shortcut edges, analogous to hopsets in graph algorithms, than as directly confronting worst-case NP-completeness barriers. Interpreting learned macros as advice/shortcuts in the Cayley graph yields a trade-off between the size of a given macro library (advice) and the achievable forward drift $\mu$: larger libraries increase $\mu$ and thus reduce expected hitting time $\approx G_n / \mu$. We conjecture that, under natural scramble distributions, practical solving is best modeled as an advice-aided average-case process: given a fixed body of learned algorithms, solvers achieve efficient performance on typical scrambles.

Exponential improvements in Rubik’s Cube performance over recent decades provide strong support for the idea that expert skill can be achieved through extensive practice \cite{Ericsson1994-zw}. Moreover, the use of algorithms to solve the cube aligns with findings from chess expertise, suggesting that pattern recognition plays a more important role than search \cite{de2008thought,connors2011expertise}. Competitive solvers simply do not have the time to search for solutions and instead search for patterns whose solutions are guaranteed by algorithms—they cache the outcome of frequently experienced computations \cite{Haith2018-om}. Moreover, competitive solvers are already operating at a near-optimal level of motor skill on the Rubik's cube. There is no evidence in either sighted or blindfolded solves for any increase in the learning rate, which is already asymptotic in the community of experts. Expertise seems to involve the formation of new representations of sets of cube configurations, akin to the ability to recall long lists of natural numbers by exploiting “sequential encoding processes and distinctive retrieval cues based on lists of pre-memorized lists of familiar physical locations” \cite{Ericsson2017-av}. The delay across decades in the acquisition of new algorithms leads to plateaus in which performance gains are slow and rapid leaps in which knowledge reaches a critical density accelerating performance \cite{Gray2017-xu}. The enormity of the Rubik’s Cube state space pushes the problem far beyond the anecdotal 10,000-hour rule, toward the possibility of effectively unbounded expertise. In this sense, a single human lifetime is not sufficient to become fully expert.

The surprising synchronous performance increases in blindfold solving in the first 8 years of competition across all cubes (3–5b-cubes) provide indirect evidence for the coordinated acquisition of new forms of skilled memory storage and acquisition. These are all based on increasingly elaborate “memory palaces”, starting with simple linear routes and evolving into filtered multi-layer palaces accompanied by the use of audio loops for edges (there are many online resources facilitating this knowledge-sharing, e.g., \cite{artofmemory-forum,magnetic-memory-palace}).

Collective intelligence is distinguished from individual intelligence by the alignment of individual skill and expertise with group-level knowledge. This represents a special case of the general explore–exploit trade-off \cite{hills2015exploration}. In collective-intelligence tasks such as those solved by social insects, this is illustrated by stigmergy \cite{detrain2008collective}, and collective decision-making through swarming \cite{tereshko2005collective}, or in large populations of cells in the mammalian brain, “dual coding theory” \cite{daniels2017dual}. In all cases, individual components independently sample or search spaces of solutions followed by communication, generating consensus by sharing solutions globally. In this way, all competitors come into possession of the same “cognitive gadgets” \cite{heyes2018cognitive}—culturally acquired algorithms of problem solving. In the competitive Rubik's cube community, knowledge-sharing is not only evident from the use of collective algorithms but by the social network of the competitive community itself (Figure~\ref{fig:f7}). The graph of sighted solvers is fully connected, implying that a subset of competitors on the 3-cube compete on all other (4–7)-cubes. The same holds for the $n_b$-cubes. This structure supports the idea that the cube defines two universality classes of problem solving with expertise shared across communities. The great interest in the human population for the Rubik's cube is that it is a cognitive artifact \cite{Norman1991} that supports the development and use of cognitive gadgets \cite{heyes2018cognitive}. The near limitless fascination with the cube is that it is not merely a puzzle but a cognitive gadget factory. 

The set of phenomena that includes cultural evolution \cite{mesoudi2016cultural}, the extended mind \cite{Clark1998-vi}, distributed cognition \cite{Zhang1994-ih}, cognitive artifacts \cite{Norman1991}, and collective intelligence \cite{centola2022network} all contend with the interactions of individual brains with social groups and material culture. Puzzles and games solved by large distributed populations connected through competitions and online platforms provide near-ideal model systems for these phenomena \cite{Chase1973,Lindstedt2020-sk,Allen2024-wu}. In this contribution, we have shown how very hard mathematical problems can be solved and reduced to universal learning classes through the adoption of shared algorithms and artifacts. These findings have implications for the understanding of collective human intelligence, which is frequently materially enhanced, and for novel forms of analog pedagogy. 

\section*{Acknowledgements}

D.C.K.\ and G.K.\ are supported by the Templeton World Charity Foundation, Inc.\ (funder DOI 501100011730) under grant DOI \href{https://doi.org/10.54224/20650}{10.54224/20650} no.\ 20650 on “Building Diverse Intelligences through Compositionality and Mechanism Design.” D.C.K.\ is additionally supported by grant no.\ 81366 from the Robert Wood Johnson Foundation on using emergent engineering for integrating complex systems to achieve an equitable society. G.K.\ is additionally supported from J.\ Grochow and R.\ Frongillo startup funds at the University of Colorado Boulder. J.A.G.\ was partially supported by NSF CAREER award CCF-2047756.

\bibliography{rubik}

\providecommand{\noopsort}[1]{}\providecommand{\singleletter}[1]{#1}

\begin{thebibliography}{39}
\ifx \bisbn   \undefined \def \bisbn  #1{ISBN #1}\fi
\ifx \binits  \undefined \def \binits#1{#1}\fi
\ifx \bauthor  \undefined \def \bauthor#1{#1}\fi
\ifx \batitle  \undefined \def \batitle#1{#1}\fi
\ifx \bjtitle  \undefined \def \bjtitle#1{#1}\fi
\ifx \bvolume  \undefined \def \bvolume#1{\textbf{#1}}\fi
\ifx \byear  \undefined \def \byear#1{#1}\fi
\ifx \bissue  \undefined \def \bissue#1{#1}\fi
\ifx \bfpage  \undefined \def \bfpage#1{#1}\fi
\ifx \blpage  \undefined \def \blpage #1{#1}\fi
\ifx \burl  \undefined \def \burl#1{\textsf{#1}}\fi
\ifx \doiurl  \undefined \def \doiurl#1{\url{https://doi.org/#1}}\fi
\ifx \betal  \undefined \def \betal{\textit{et al.}}\fi
\ifx \binstitute  \undefined \def \binstitute#1{#1}\fi
\ifx \binstitutionaled  \undefined \def \binstitutionaled#1{#1}\fi
\ifx \bctitle  \undefined \def \bctitle#1{#1}\fi
\ifx \beditor  \undefined \def \beditor#1{#1}\fi
\ifx \bpublisher  \undefined \def \bpublisher#1{#1}\fi
\ifx \bbtitle  \undefined \def \bbtitle#1{#1}\fi
\ifx \bedition  \undefined \def \bedition#1{#1}\fi
\ifx \bseriesno  \undefined \def \bseriesno#1{#1}\fi
\ifx \blocation  \undefined \def \blocation#1{#1}\fi
\ifx \bsertitle  \undefined \def \bsertitle#1{#1}\fi
\ifx \bsnm \undefined \def \bsnm#1{#1}\fi
\ifx \bsuffix \undefined \def \bsuffix#1{#1}\fi
\ifx \bparticle \undefined \def \bparticle#1{#1}\fi
\ifx \barticle \undefined \def \barticle#1{#1}\fi
\bibcommenthead
\ifx \bconfdate \undefined \def \bconfdate #1{#1}\fi
\ifx \botherref \undefined \def \botherref #1{#1}\fi
\ifx \url \undefined \def \url#1{\textsf{#1}}\fi
\ifx \bchapter \undefined \def \bchapter#1{#1}\fi
\ifx \bbook \undefined \def \bbook#1{#1}\fi
\ifx \bcomment \undefined \def \bcomment#1{#1}\fi
\ifx \oauthor \undefined \def \oauthor#1{#1}\fi
\ifx \citeauthoryear \undefined \def \citeauthoryear#1{#1}\fi
\ifx \endbibitem  \undefined \def \endbibitem {}\fi
\ifx \bconflocation  \undefined \def \bconflocation#1{#1}\fi
\ifx \arxivurl  \undefined \def \arxivurl#1{\textsf{#1}}\fi
\csname PreBibitemsHook\endcsname

\bibitem[\protect\citeauthoryear{Cooperman et~al.}{1991}]{cooperman1990applications}
\begin{bchapter}
\bauthor{\bsnm{Cooperman}, \binits{G.}},
\bauthor{\bsnm{Finkelstein}, \binits{L.}},
\bauthor{\bsnm{Sarawagi}, \binits{N.}}:
\bctitle{Applications of {Cayley} graphs}.
In: \bbtitle{Appl. Algebra, Algebraic Algorithms \& Error-Correcting Codes: 8th Internat. Conf. (AAECC-8, '90)},
pp. \bfpage{367}--\blpage{378}
(\byear{1991}).
\doiurl{10.1007/3-540-54195-0_65} .
\bcomment{Springer}
\end{bchapter}
\endbibitem

\bibitem[\protect\citeauthoryear{Rokicki}{2020}]{rokicki2014thirty}
\begin{bchapter}
\bauthor{\bsnm{Rokicki}, \binits{T.}}:
\bctitle{Thirty years of computer cubing: The search for {God's} number}.
In: \beditor{\bsnm{Plambeck}, \binits{T.}},
\beditor{\bsnm{Rokicki}, \binits{T.}} (eds.)
\bbtitle{Barrycades and Septoku: Papers in Honor of Martin Gardner and Tom Rogers},
pp. \bfpage{79}--\blpage{98}.
\bpublisher{MAA Press},
\blocation{USA}
(\byear{2020}).
\bcomment{Reprint; originally published 2014.}
\end{bchapter}
\endbibitem

\bibitem[\protect\citeauthoryear{Rothstein}{2019}]{rothstein2019shape}
\begin{bbook}
\bauthor{\bsnm{Rothstein}, \binits{B.L.}}:
\bbtitle{The Shape of Difficulty: A Fan Letter to Unruly Objects}.
\bpublisher{Pennsylvania State University Press},
\blocation{University Park, PA, USA}
(\byear{2019})
\end{bbook}
\endbibitem

\bibitem[\protect\citeauthoryear{Danesi}{2020}]{danesi2020anthropology}
\begin{botherref}
\oauthor{\bsnm{Danesi}, \binits{M.}}:
An anthropology of puzzles: The role of puzzles in the origins and evolution of mind and culture.
Routledge
(2020).
\doiurl{10.4324/9781003084495}
\end{botherref}
\endbibitem

\bibitem[\protect\citeauthoryear{Karde\c{s} et~al.}{2025}]{KardesKrakauerGrochow2025PCoCA}
\begin{botherref}
\oauthor{\bsnm{Karde\c{s}}, \binits{G.}},
\oauthor{\bsnm{Krakauer}, \binits{D.C.}},
\oauthor{\bsnm{Grochow}, \binits{J.A.}}:
Physical Complexity of a Cognitive Artifact.
Manuscript, preprint \href{https://arxiv.org/abs/2509.12495}{arXiv:2509.12495}
(2025)
\end{botherref}
\endbibitem

\bibitem[\protect\citeauthoryear{Scheffler}{2017}]{scheffler2017cracking}
\begin{bbook}
\bauthor{\bsnm{Scheffler}, \binits{I.}}:
\bbtitle{Cracking the Cube: Going Slow to Go Fast and Other Unexpected Turns in the World of Competitive Rubik's Cube Solving}.
\bpublisher{Simon and Schuster},
\blocation{USA}
(\byear{2017})
\end{bbook}
\endbibitem

\bibitem[\protect\citeauthoryear{Harris}{2008}]{harris2008speedsolving}
\begin{bbook}
\bauthor{\bsnm{Harris}, \binits{D.}}:
\bbtitle{Speedsolving the Cube}.
\bpublisher{Sterling Publishing Company, Inc.},
\blocation{USA}
(\byear{2008})
\end{bbook}
\endbibitem

\bibitem[\protect\citeauthoryear{Duberg and Tidestr{\"o}m}{2015}]{duberg2015comparison}
\begin{botherref}
\oauthor{\bsnm{Duberg}, \binits{D.}},
\oauthor{\bsnm{Tidestr{\"o}m}, \binits{J.}}:
Comparison of rubik’s cube solving methods made for humans.
Degree project, first level
(2015).
\href{https://urn.kb.se/resolve?urn=urn%3Anbn%3Ase%3Akth%3Adiva-166727}{ urn:nbn:se:kth:diva-166727}
\end{botherref}
\endbibitem

\bibitem[\protect\citeauthoryear{Rubik}{2024}]{rubik2024rubik}
\begin{bbook}
\bauthor{\bsnm{Rubik}, \binits{E.}}:
\bbtitle{Rubik's: 50 Years of the World's Most Famous Cube}.
\bpublisher{White Lion},
\blocation{USA}
(\byear{2024})
\end{bbook}
\endbibitem

\bibitem[\protect\citeauthoryear{Yates}{2013}]{yates2013art}
\begin{bbook}
\bauthor{\bsnm{Yates}, \binits{F.A.}}:
\bbtitle{The Art of Memory}.
\bpublisher{Routledge},
\blocation{UK}
(\byear{2013})
\end{bbook}
\endbibitem

\bibitem[\protect\citeauthoryear{Carruthers}{1992}]{carruthers1992book}
\begin{bbook}
\bauthor{\bsnm{Carruthers}, \binits{M.J.}}:
\bbtitle{The Book of Memory: A Study of Memory in Medieval Culture}
vol. \bseriesno{10}.
\bpublisher{Cambridge University Press},
\blocation{UK}
(\byear{1992})
\end{bbook}
\endbibitem

\bibitem[\protect\citeauthoryear{Murre}{2014}]{murre2014s}
\begin{barticle}
\bauthor{\bsnm{Murre}, \binits{J.M.}}:
\batitle{S-shaped learning curves}.
\bjtitle{Psychonomic Bulletin \& Review}
\bvolume{21}(\bissue{2}),
\bfpage{344}--\blpage{356}
(\byear{2014})
\doiurl{10.3758/s13423-013-0522-0}
\end{barticle}
\endbibitem

\bibitem[\protect\citeauthoryear{Sun et~al.}{2025}]{sun2024snseg}
\begin{barticle}
\bauthor{\bsnm{Sun}, \binits{S.}},
\bauthor{\bsnm{Zhao}, \binits{Z.}},
\bauthor{\bsnm{Jiang}, \binits{F.}},
\bauthor{\bsnm{Shao}, \binits{X.}}:
\batitle{Snseg: An {R} package for time series segmentation via self-normalization}.
\bjtitle{The R Journal}
(\byear{2025})
\doiurl{10.32614/RJ-2024-029}
\end{barticle}
\endbibitem

\bibitem[\protect\citeauthoryear{Newman}{2006}]{newman2006modularity}
\begin{barticle}
\bauthor{\bsnm{Newman}, \binits{M.E.J.}}:
\batitle{Modularity and community structure in networks}.
\bjtitle{Proceedings of the National Academy of Sciences}
\bvolume{103}(\bissue{23}),
\bfpage{8577}--\blpage{8582}
(\byear{2006})
\doiurl{10.1073/pnas.0601602103}
\end{barticle}
\endbibitem

\bibitem[\protect\citeauthoryear{Demaine et~al.}{2018}]{demaine2017solving}
\begin{bchapter}
\bauthor{\bsnm{Demaine}, \binits{E.D.}},
\bauthor{\bsnm{Eisenstat}, \binits{S.}},
\bauthor{\bsnm{Rudoy}, \binits{M.}}:
\bctitle{Solving the {Rubik's} cube optimally is {NP}-complete}.
In: \beditor{\bsnm{Niedermeier}, \binits{R.}},
\beditor{\bsnm{Vall\'{e}e}, \binits{B.}} (eds.)
\bbtitle{35th Symposium on Theoretical Aspects of Computer Science (STACS 2018)}.
\bsertitle{Leibniz International Proceedings in Informatics (LIPIcs)},
vol. \bseriesno{96},
pp. \bfpage{24}--\blpage{12413}.
\bpublisher{Schloss Dagstuhl -- Leibniz-Zentrum f{\"u}r Informatik},
\blocation{Dagstuhl, Germany}
(\byear{2018}).
\doiurl{10.4230/LIPIcs.STACS.2018.24} .
\bcomment{arXiv preprint \href{https://arxiv.org/abs/1706.06708}{arXiv:1706.06708}}
\end{bchapter}
\endbibitem

\bibitem[\protect\citeauthoryear{Brush et~al.}{2018}]{brush2018conflicts}
\begin{barticle}
\bauthor{\bsnm{Brush}, \binits{E.R.}},
\bauthor{\bsnm{Krakauer}, \binits{D.C.}},
\bauthor{\bsnm{Flack}, \binits{J.C.}}:
\batitle{Conflicts of interest improve collective computation of adaptive social structures}.
\bjtitle{Science Advances}
\bvolume{4}(\bissue{1}),
\bfpage{1603311}
(\byear{2018})
\doiurl{10.1126/sciadv.1603311}
\end{barticle}
\endbibitem

\bibitem[\protect\citeauthoryear{Hutchins}{2014}]{Hutchins2014-qa}
\begin{bbook}
\bauthor{\bsnm{Hutchins}, \binits{E.}}:
\bbtitle{Cognition in the Wild}.
\bpublisher{MIT Press},
\blocation{Cambridge, MA, USA}
(\byear{2014}).
\doiurl{10.7551/mitpress/1881.001.0001}
\end{bbook}
\endbibitem

\bibitem[\protect\citeauthoryear{Clark and Chalmers}{1998}]{Clark1998-vi}
\begin{barticle}
\bauthor{\bsnm{Clark}, \binits{A.}},
\bauthor{\bsnm{Chalmers}, \binits{D.}}:
\batitle{The extended mind}.
\bjtitle{Analysis}
\bvolume{58}(\bissue{1}),
\bfpage{7}--\blpage{19}
(\byear{1998})
\doiurl{10.1093/analys/58.1.7}
\end{barticle}
\endbibitem

\bibitem[\protect\citeauthoryear{Krakauer}{2016}]{krakauer2016will}
\begin{botherref}
\oauthor{\bsnm{Krakauer}, \binits{D.}}:
Will {AI} harm us? {Better} to ask how we’ll reckon with our hybrid nature.
Nautilus
\textbf{6}
(2016)
\end{botherref}
\endbibitem

\bibitem[\protect\citeauthoryear{Ericsson and Charness}{1994}]{Ericsson1994-zw}
\begin{barticle}
\bauthor{\bsnm{Ericsson}, \binits{K.A.}},
\bauthor{\bsnm{Charness}, \binits{N.}}:
\batitle{Expert performance: Its structure and acquisition}.
\bjtitle{Am. Psychol.}
\bvolume{49}(\bissue{8}),
\bfpage{725}--\blpage{747}
(\byear{1994})
\doiurl{10.10370003-066X.49.8.725}
\end{barticle}
\endbibitem

\bibitem[\protect\citeauthoryear{De~Groot}{2008}]{de2008thought}
\begin{bbook}
\bauthor{\bsnm{De~Groot}, \binits{A.D.}}:
\bbtitle{Thought and Choice in Chess}.
\bpublisher{Amsterdam University Press}, \blocation{???}
(\byear{2008})
\end{bbook}
\endbibitem

\bibitem[\protect\citeauthoryear{Connors et~al.}{2011}]{connors2011expertise}
\begin{barticle}
\bauthor{\bsnm{Connors}, \binits{M.H.}},
\bauthor{\bsnm{Burns}, \binits{B.D.}},
\bauthor{\bsnm{Campitelli}, \binits{G.}}:
\batitle{Expertise in complex decision making: The role of search in chess 70 years after de {Groot}}.
\bjtitle{Cognitive Science}
\bvolume{35}(\bissue{8}),
\bfpage{1567}--\blpage{1579}
(\byear{2011})
\doiurl{10.1111/j.1551-6709.2011.01196.x}
\end{barticle}
\endbibitem

\bibitem[\protect\citeauthoryear{Haith and Krakauer}{2018}]{Haith2018-om}
\begin{barticle}
\bauthor{\bsnm{Haith}, \binits{A.M.}},
\bauthor{\bsnm{Krakauer}, \binits{J.W.}}:
\batitle{The multiple effects of practice: skill, habit and reduced cognitive load}.
\bjtitle{Curr. Opin. Behav. Sci.}
\bvolume{20},
\bfpage{196}--\blpage{201}
(\byear{2018})
\doiurl{10.1016/j.cobeha.2018.01.015}
\end{barticle}
\endbibitem

\bibitem[\protect\citeauthoryear{Ericsson et~al.}{2017}]{Ericsson2017-av}
\begin{barticle}
\bauthor{\bsnm{Ericsson}, \binits{K.A.}},
\bauthor{\bsnm{Cheng}, \binits{X.}},
\bauthor{\bsnm{Pan}, \binits{Y.}},
\bauthor{\bsnm{Ku}, \binits{Y.}},
\bauthor{\bsnm{Ge}, \binits{Y.}},
\bauthor{\bsnm{Hu}, \binits{Y.}}:
\batitle{Memory skills mediating superior memory in a world-class memorist}.
\bjtitle{Memory}
\bvolume{25}(\bissue{9}),
\bfpage{1294}--\blpage{1302}
(\byear{2017})
\doiurl{10.1080/09658211.2017.1296164}
\end{barticle}
\endbibitem

\bibitem[\protect\citeauthoryear{Gray and Lindstedt}{2017}]{Gray2017-xu}
\begin{barticle}
\bauthor{\bsnm{Gray}, \binits{W.D.}},
\bauthor{\bsnm{Lindstedt}, \binits{J.K.}}:
\batitle{Plateaus, dips, and leaps: Where to look for inventions and discoveries during skilled performance}.
\bjtitle{Cogn. Sci.}
\bvolume{41}(\bissue{7}),
\bfpage{1838}--\blpage{1870}
(\byear{2017})
\doiurl{10.1111/cogs.12412}
\end{barticle}
\endbibitem

\bibitem[\protect\citeauthoryear{}{}]{artofmemory-forum}
\begin{botherref}
Art of Memory Forum.
\url{https://forum.artofmemory.com/}.
Accessed: 2025-08-30
\end{botherref}
\endbibitem

\bibitem[\protect\citeauthoryear{{Magnetic Memory Method}}{}]{magnetic-memory-palace}
\begin{botherref}
\oauthor{\bsnm{{Magnetic Memory Method}}}:
Memory Palace.
\url{https://www.magneticmemorymethod.com/memory-palace/}.
Accessed: 2025-08-30
\end{botherref}
\endbibitem

\bibitem[\protect\citeauthoryear{Hills et~al.}{2015}]{hills2015exploration}
\begin{barticle}
\bauthor{\bsnm{Hills}, \binits{T.T.}},
\bauthor{\bsnm{Todd}, \binits{P.M.}},
\bauthor{\bsnm{Lazer}, \binits{D.}},
\bauthor{\bsnm{Redish}, \binits{A.D.}},
\bauthor{\bsnm{Couzin}, \binits{I.D.}}:
\batitle{Exploration versus exploitation in space, mind, and society}.
\bjtitle{Trends in Cognitive Sciences}
\bvolume{19}(\bissue{1}),
\bfpage{46}--\blpage{54}
(\byear{2015})
\doiurl{10.1016/j.tics.2014.10.004}
\end{barticle}
\endbibitem

\bibitem[\protect\citeauthoryear{Detrain and Deneubourg}{2008}]{detrain2008collective}
\begin{barticle}
\bauthor{\bsnm{Detrain}, \binits{C.}},
\bauthor{\bsnm{Deneubourg}, \binits{J.-L.}}:
\batitle{Collective decision-making and foraging patterns in ants and honeybees}.
\bjtitle{Advances in Insect Physiology}
\bvolume{35},
\bfpage{123}--\blpage{173}
(\byear{2008})
\doiurl{10.1016/S0065-2806(08)00002-7}
\end{barticle}
\endbibitem

\bibitem[\protect\citeauthoryear{Tereshko and Loengarov}{2005}]{tereshko2005collective}
\begin{barticle}
\bauthor{\bsnm{Tereshko}, \binits{V.}},
\bauthor{\bsnm{Loengarov}, \binits{A.}}:
\batitle{Collective decision making in honey-bee foraging dynamics}.
\bjtitle{Computing and Information Systems}
\bvolume{9}(\bissue{3}),
\bfpage{1}
(\byear{2005})
\end{barticle}
\endbibitem

\bibitem[\protect\citeauthoryear{Daniels et~al.}{2017}]{daniels2017dual}
\begin{barticle}
\bauthor{\bsnm{Daniels}, \binits{B.C.}},
\bauthor{\bsnm{Flack}, \binits{J.C.}},
\bauthor{\bsnm{Krakauer}, \binits{D.C.}}:
\batitle{Dual coding theory explains biphasic collective computation in neural decision-making}.
\bjtitle{Frontiers in Neuroscience}
\bvolume{11},
\bfpage{313}
(\byear{2017})
\doiurl{10.3389/fnins.2017.00313}
\end{barticle}
\endbibitem

\bibitem[\protect\citeauthoryear{Heyes}{2018}]{heyes2018cognitive}
\begin{bbook}
\bauthor{\bsnm{Heyes}, \binits{C.}}:
\bbtitle{Cognitive Gadgets: The Cultural Evolution of Thinking}.
\bpublisher{Harvard University Press},
\blocation{Cambridge, MA, USA}
(\byear{2018})
\end{bbook}
\endbibitem

\bibitem[\protect\citeauthoryear{Norman}{1991}]{Norman1991}
\begin{bchapter}
\bauthor{\bsnm{Norman}, \binits{D.A.}}:
\bctitle{Cognitive artifacts}.
In: \beditor{\bsnm{Carroll}, \binits{J.M.}} (ed.)
\bbtitle{Designing Interaction: Psychology at the Human-computer Interface},
pp. \bfpage{17}--\blpage{38}.
\bpublisher{Cambridge University Press},
\blocation{UK}
(\byear{1991})
\end{bchapter}
\endbibitem

\bibitem[\protect\citeauthoryear{Mesoudi}{2016}]{mesoudi2016cultural}
\begin{barticle}
\bauthor{\bsnm{Mesoudi}, \binits{A.}}:
\batitle{Cultural evolution: a review of theory, findings and controversies}.
\bjtitle{Evolutionary Biology}
\bvolume{43}(\bissue{4}),
\bfpage{481}--\blpage{497}
(\byear{2016})
\doiurl{10.1007/s11692-015-9320-0}
\end{barticle}
\endbibitem

\bibitem[\protect\citeauthoryear{Zhang and Norman}{1994}]{Zhang1994-ih}
\begin{barticle}
\bauthor{\bsnm{Zhang}, \binits{J.}},
\bauthor{\bsnm{Norman}, \binits{D.A.}}:
\batitle{Representations in distributed cognitive tasks}.
\bjtitle{Cogn. Sci.}
\bvolume{18}(\bissue{1}),
\bfpage{87}--\blpage{122}
(\byear{1994})
\doiurl{10.1016/0364-0213(94)90021-3}
\end{barticle}
\endbibitem

\bibitem[\protect\citeauthoryear{Centola}{2022}]{centola2022network}
\begin{barticle}
\bauthor{\bsnm{Centola}, \binits{D.}}:
\batitle{The network science of collective intelligence}.
\bjtitle{Trends in Cognitive Sciences}
\bvolume{26}(\bissue{11}),
\bfpage{923}--\blpage{941}
(\byear{2022})
\doiurl{10.1016/j.tics.2022.08.009}
\end{barticle}
\endbibitem

\bibitem[\protect\citeauthoryear{Chase and Simon}{1973}]{Chase1973}
\begin{barticle}
\bauthor{\bsnm{Chase}, \binits{W.G.}},
\bauthor{\bsnm{Simon}, \binits{H.A.}}:
\batitle{Perception in chess}.
\bjtitle{Cognitive Psychology}
\bvolume{4}(\bissue{1}),
\bfpage{55}--\blpage{81}
(\byear{1973})
\doiurl{10.1016/0010-0285(73)90004-2}
\end{barticle}
\endbibitem

\bibitem[\protect\citeauthoryear{Lindstedt and Gray}{2020}]{Lindstedt2020-sk}
\begin{bchapter}
\bauthor{\bsnm{Lindstedt}, \binits{J.K.}},
\bauthor{\bsnm{Gray}, \binits{W.D.}}:
\bctitle{The ``cognitive speed-bump''': How world champion {Tetris} players trade milliseconds for seconds}.
In: \beditor{\bsnm{Denison}, \binits{S.}},
\beditor{\bsnm{Mack}, \binits{M.L.}},
\beditor{\bsnm{Xu}, \binits{Y.}},
\beditor{\bsnm{Armstrong}, \binits{B.C.}} (eds.)
\bbtitle{Proc. 42nd Annual Meeting of the Cog. Sci. Soc. (CogSci 2020)}
(\byear{2020}).
\burl{https://cognitivesciencesociety.org/cogsci-2020/}
\end{bchapter}
\endbibitem

\bibitem[\protect\citeauthoryear{Allen et~al.}{2024}]{Allen2024-wu}
\begin{barticle}
\bauthor{\bsnm{Allen}, \binits{K.}},
\bauthor{\bsnm{Brändle}, \binits{F.}},
\bauthor{\bsnm{Botvinick}, \binits{M.}},
\bauthor{\bsnm{Fan}, \binits{J.E.}},
\bauthor{\bsnm{Gershman}, \binits{S.J.}},
\bauthor{\bsnm{Gopnik}, \binits{A.}},
\bauthor{\bsnm{Griffiths}, \binits{T.L.}},
\bauthor{\bsnm{Hartshorne}, \binits{J.K.}},
\bauthor{\bsnm{Hauser}, \binits{T.U.}},
\bauthor{\bsnm{Ho}, \binits{M.K.}},
\bauthor{\bsnm{Leeuw}, \binits{J.R.}},
\bauthor{\bsnm{Ma}, \binits{W.J.}},
\bauthor{\bsnm{Murayama}, \binits{K.}},
\bauthor{\bsnm{Nelson}, \binits{J.D.}},
\bauthor{\bsnm{Opheusden}, \binits{B.}},
\bauthor{\bsnm{Pouncy}, \binits{T.}},
\bauthor{\bsnm{Rafner}, \binits{J.}},
\bauthor{\bsnm{Rahwan}, \binits{I.}},
\bauthor{\bsnm{Rutledge}, \binits{R.B.}},
\bauthor{\bsnm{Sherson}, \binits{J.}},
\bauthor{\bsnm{Simsek}, \binits{O.}},
\bauthor{\bsnm{Spiers}, \binits{H.}},
\bauthor{\bsnm{Summerfield}, \binits{C.}},
\bauthor{\bsnm{Thalmann}, \binits{M.}},
\bauthor{\bsnm{Vélez}, \binits{N.}},
\bauthor{\bsnm{Watrous}, \binits{A.J.}},
\bauthor{\bsnm{Tenenbaum}, \binits{J.B.}},
\bauthor{\bsnm{Schulz}, \binits{E.}}:
\batitle{Using games to understand the mind}.
\bjtitle{Nat. Hum. Behav.}
\bvolume{8}(\bissue{6}),
\bfpage{1035}--\blpage{1043}
(\byear{2024})
\doiurl{10.1038/s41562-024-01878-9}
\end{barticle}
\endbibitem

\end{thebibliography}

\end{document}